\documentclass[conference]{IEEEtran}
\IEEEoverridecommandlockouts
\usepackage{cite}
\usepackage{amsmath,amssymb,amsfonts}
\usepackage{algorithmic}
\usepackage{graphicx}
\usepackage{algorithm}
\usepackage{textcomp}
\usepackage{xcolor}
\def\BibTeX{{\rm B\kern-.05em{\sc i\kern-.025em b}\kern-.08em
    T\kern-.1667em\lower.7ex\hbox{E}\kern-.125emX}}

\usepackage[utf8]{inputenc} 
\usepackage[T1]{fontenc}    
\usepackage{url}            
\usepackage{booktabs}       
\usepackage{nicefrac}       
\usepackage{microtype}      
\usepackage{makecell}
\usepackage{enumitem}
\usepackage{multicol,multirow}
\newcommand{\INPUT}{\item[\textbf{Input:}]}
\newcommand{\OUTPUT}{\item[\textbf{Output:}]}
\usepackage{wrapfig}

\begin{document}


\title{First-order State Space Model for Lightweight Image Super-resolution}

\author{

\IEEEauthorblockN{
Yujie Zhu\textsuperscript{1},
Xinyi Zhang\textsuperscript{1},
Yekai Lu\textsuperscript{1},
Guang Yang\textsuperscript{1,2},
Faming Fang\textsuperscript{1},
Guixu Zhang\textsuperscript{1}\IEEEauthorrefmark{1}\thanks{\IEEEauthorrefmark{1}Corresponding author}
\thanks{This work was supported by the National Key R\&D Program of China (2022ZD0161800),
and the National Natural Science Foundation of China under Grant 62271203.}
}

\IEEEauthorblockA{
\textsuperscript{1}School of Computer Science and Technology, East China Normal University, Shanghai, China\\
\textsuperscript{2}IT Department, Guotai Junan Security, Shanghai, China\\
52205901006@stu.ecnu.edu.cn, 
51265901099@stu.ecnu.edu.cn,
51255901081@stu.ecnu.edu.cn, \\
51215901104@stu.ecnu.edu.cn,
fmfang@cs.ecnu.edu.cn,
gxzhang@cs.ecnu.edu.cn
}
}
\IEEEaftertitletext{\vspace{-1\baselineskip}}
\maketitle
\begin{abstract}


State space models (SSMs), particularly Mamba, have shown promise in NLP tasks and are increasingly applied to vision tasks. However, most Mamba-based vision models focus on network architecture and scan paths, with little attention to the SSM module. In order to explore the potential of SSMs, we modified the calculation process of SSM without increasing the number of parameters to improve the performance on lightweight super-resolution tasks. In this paper, we introduce the First-order State Space Model (FSSM) to improve the original Mamba module, enhancing performance by incorporating token correlations. We apply a first-order hold condition in SSMs, derive the new discretized form, and analyzed cumulative error. Extensive experimental results demonstrate that FSSM improves the performance of MambaIR on five benchmark datasets without additionally increasing the number of parameters, and surpasses current lightweight SR methods, achieving state-of-the-art results. Code is available at https://github.com/Edlinf/FMambaIR.

\end{abstract}


\section{Introduction}
Single image super-resolution (SR) aims to reconstruct a high-resolution output image from a given low-resolution image and is a well-known problem in computer vision. The development of deep learning models has significantly improved SR performance in recent years. Among these models, CNNs and Transformers are now dominant. CNNs efficiently combine information from adjacent pixels, resulting in lower computation costs but limiting their ability to capture global image information due to their restricted receptive fields. In contrast, Transformers can handle long-range dependencies between pixels, expanding their receptive fields but increasing computation cost, especially at higher resolutions.

The size of the receptive field greatly influences the representation capability of neural networks. However, balancing low computation cost with a large receptive field is challenging. Recently, structured state-space sequence (S4) models\cite{gu2021efficiently}, particularly Mamba\cite{gu2023mamba}, have provided a way to balance global receptive fields with manageable computational costs. Mamba’s recursive state-space equations model long-range dependencies with linear complexity. Additionally, Hippo\cite{gu2020hippo} and H3\cite{fu2022hungry} models enhance this capability, reducing the performance gap between SSMs and Transformers. Furthermore, the parallel scan algorithm\cite{smith2022simplified,gu2023mamba} accelerates the training process of Mamba by leveraging modern GPU capabilities.

\begin{figure}[]
\begin{center}
\includegraphics[width=0.8\linewidth]{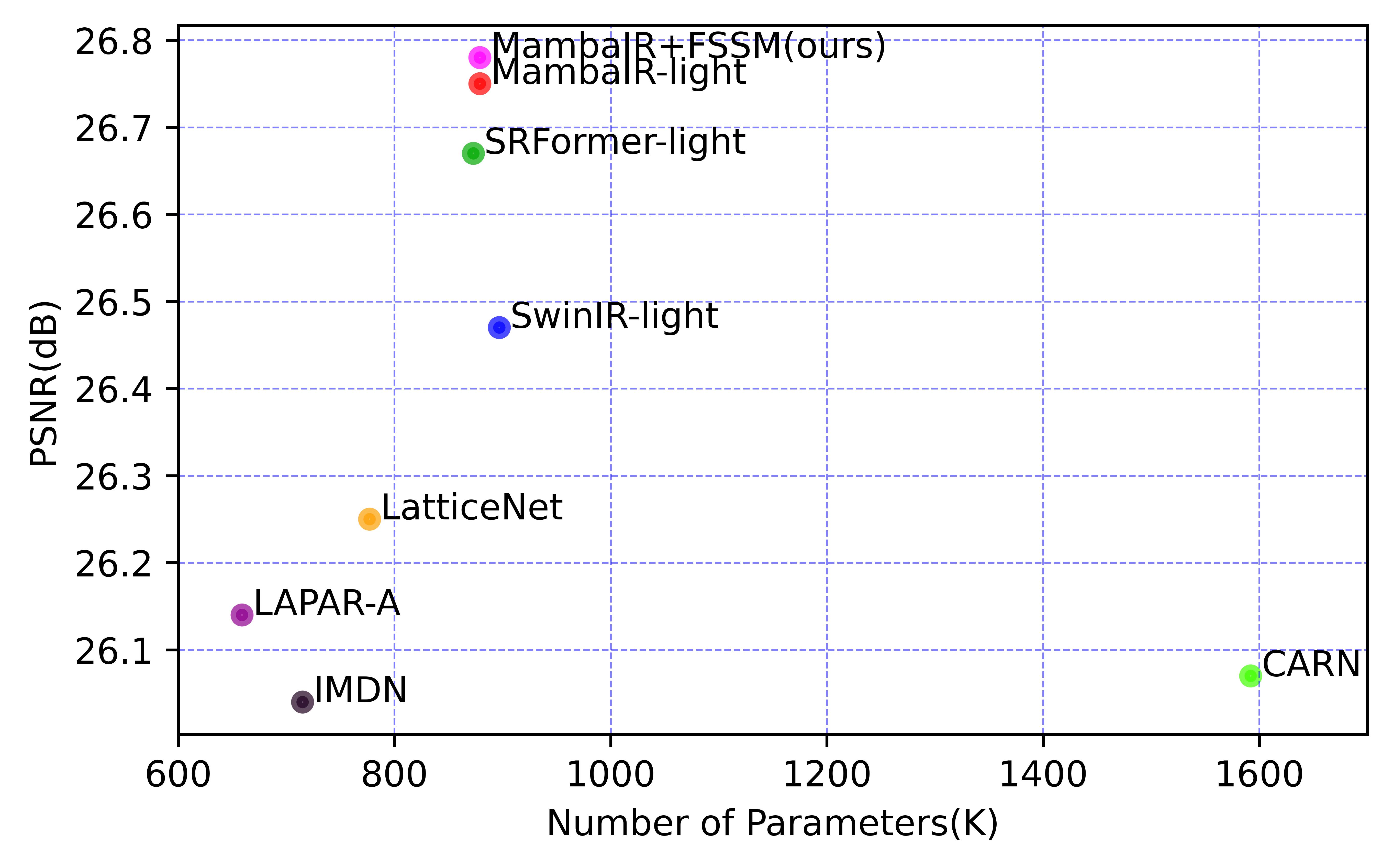}
\end{center}
\vspace{-\baselineskip}
\caption{Trade-off between the number of model parameters and performance on Urban100(x4).}
\label{urbanx4}
\vspace{-\baselineskip}
\end{figure}

Given Mamba's promising features, we introduce First-order State Space Model (FSSM), a first-order state-space model. FSSM is designed for tasks where the entire input sequence is available beforehand, enabling the use of correlations between adjacent tokens. To improve SSM module, we derive the discrete form with first-order hold condition. Besides, we analyzed the cumulative error and applied various expansion for better approximate. For enhanced computational efficiency, we also modify Mamba’s CUDA code in our FSSM implementation.

Our main contributions can be summarized as follows:
\begin{itemize}[leftmargin=*]
    \item We propose First-order State Space Model (FSSM), which replaces traditional SSM module in Mamba block, improving its representation ability. Based on first-order hold condition, we re-derive discretization form of SSM.
    \item To balance the computational efficiency and precision, we applied multiple forms of approximation, and analyzed the cumulative error.
    \item We conduct experiments on light-weight SR by replacing SSM module with our FSSM module in MambaIR~\cite{guo2024mambair}. Fig.~\ref{urbanx4} provides a visual performance comparison, which demonstrate that our FSSM module improves the performance of MambaIR, and outperforms other strong baselines on light-weight image super-resolution task.
\end{itemize}

\begin{figure}[h]
\begin{center}
\includegraphics[width=0.98\linewidth]{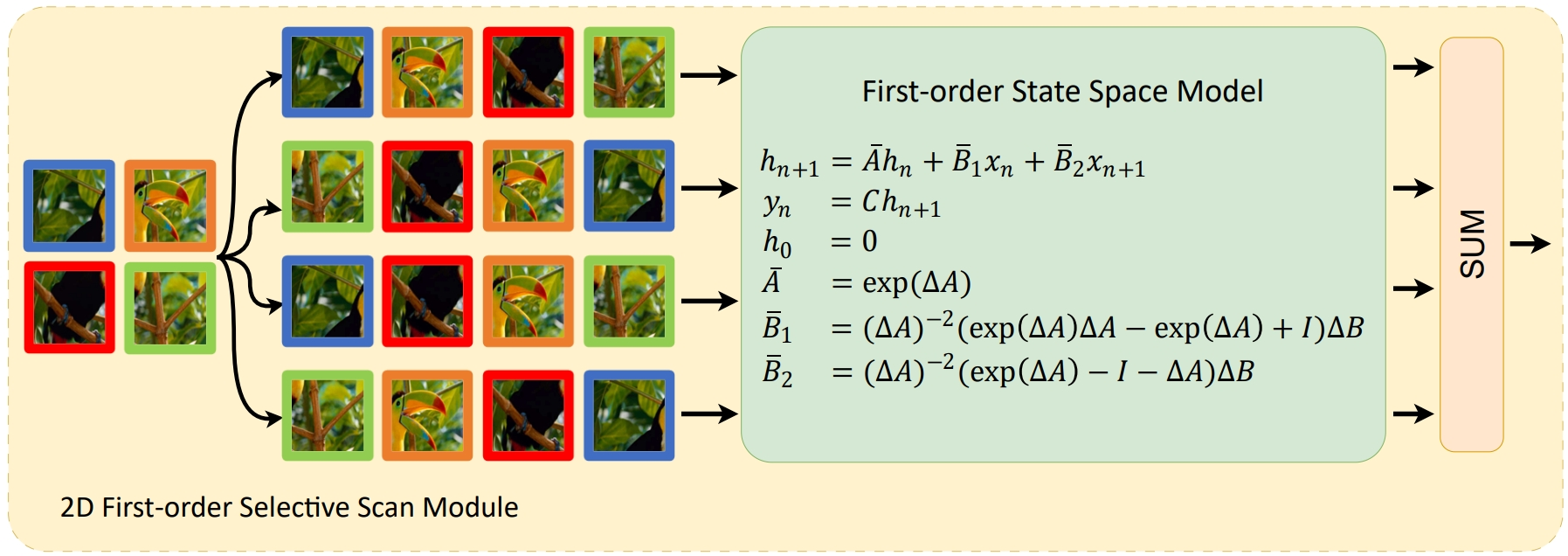}
\end{center}
\vspace{-\baselineskip}
\caption{The architecture diagram of 2D First-order Selective Scan Module.}
\label{2dfssm}
\vspace{-\baselineskip}
\end{figure}

\section{Related Works}

\subsection{Image Super-resolution}

Image super-resolution has seen significant advancements since SRCNN\cite{dong2014learning} introduced deep convolutional neural networks (CNNs) to the task. Following SRCNN, various CNN blocks like residual\cite{kim2016accurate,lim2017enhanced} and dense blocks\cite{wang2018esrgan,zhang2018residual} have been developed to enhance model representation. However, CNN-based models struggle with capturing long-range and global dependencies effectively. On the other hand, Transformer-based models have shown strong results in SR tasks, setting new benchmarks recently. Transformers excel at capturing global information but face the challenge of quadratic computational complexity due to self-attention [12]. To mitigate this, some methods, like Swin-Transformer\cite{liu2021swin} and NAT\cite{hassani2023neighborhood}, limit attention in neighboring patches or pixels. However, these approaches still face a trade-off between expanding the effective global receptive field and maintaining efficient attention computation which is a challenging issue in Transformer architecture.

\subsection{State Space Model}
Recently, the Structured State-Space Sequence (S4) model has emerged as a promising alternative to CNNs and Transformers due to its linear scaling with sequence length. The S5 layer\cite{smith2022simplified}, introduced with MIMO SSM and efficient parallel scanning, significantly accelerates training. H3\cite{fu2022hungry} further enhances the long-range dependency capability of SSMs, narrowing the performance gap with Transformers. More recently, Mamba\cite{gu2023mamba}, a data-dependent SSM, has outperformed Transformers in natural language processing while maintaining linear scaling.
To adapt Mamba for vision tasks, various methods have been proposed, including the Cross-Scan Module\cite{liu2024vmamba} and bidirectional SSM with position embeddings\cite{zhu2024vision}. Mamba has been applied to a variety of vision tasks, including image classification\cite{liu2024vmamba,zhu2024vision}, medical image segmentation\cite{ma2024u}, image generation\cite{fu2024md,hu2024zigma}, image restoration\cite{guo2024mambair,zheng2024u}, video object segmentation\cite{yang2024vivim} and others\cite{nguyen2022s4nd,islam2023efficient,liang2024pointmamba}. Our FSSM further enhances Mamba, improving it as a better backbone for super-resolution task.

\section{Method}



\subsection{Overview of 2D First-order Selective Scan Module }
We propose the 2D First-order Selective Scan Module for convenient plug-and-play. The architecture of our proposed module is shown in Fig.~\ref{2dfssm}, aims to enhance the implicit expression ability to improve the effectiveness of SSM in lightweight SR tasks. The 2D First-order Selective Scan Module inputs the features into First-order State Space Model (FSSM) along four directions, and combined the results for output. Later we will describe the details of FSSM and analyze the cumulative error.

\subsection{First-order State Space Model (FSSM)}

The classical State Space Model (SSM) is generally considered to be a linear system
\begin{equation}
\begin{aligned}
h'(t)&=A(t)h(t)+B(t)x(t) \\
y(t)&=C(t)h(t)
\end{aligned}
\label{ssm1}
\end{equation}
that maps continuous input $x(t)$ to output $y(t)$ by introducing a state variable $h(t)$, where $x(t),y(t)\in C^0(\mathbb{R})$, $B(t),C(t)\in C^0(\mathbb{R}^{D})$, and $A(t)\in C^0(\mathbb{R}^{D\times D})$.  
If the parameters of the system do not change over time $t$, it is referred to as a time-invariant SSM. For discrete data, we usually expand the discrete input $\{x_n\}$ to continuous $x(t)$, and then solve Eq.\eqref{ssm1} to get the solution $y(t)$, finally taking time index of $y(t)$ to get discrete output $\{y_n\}$.

\begin{figure}[htbp]
\vspace{-\baselineskip}
\begin{center}
\includegraphics[width=0.98\linewidth]{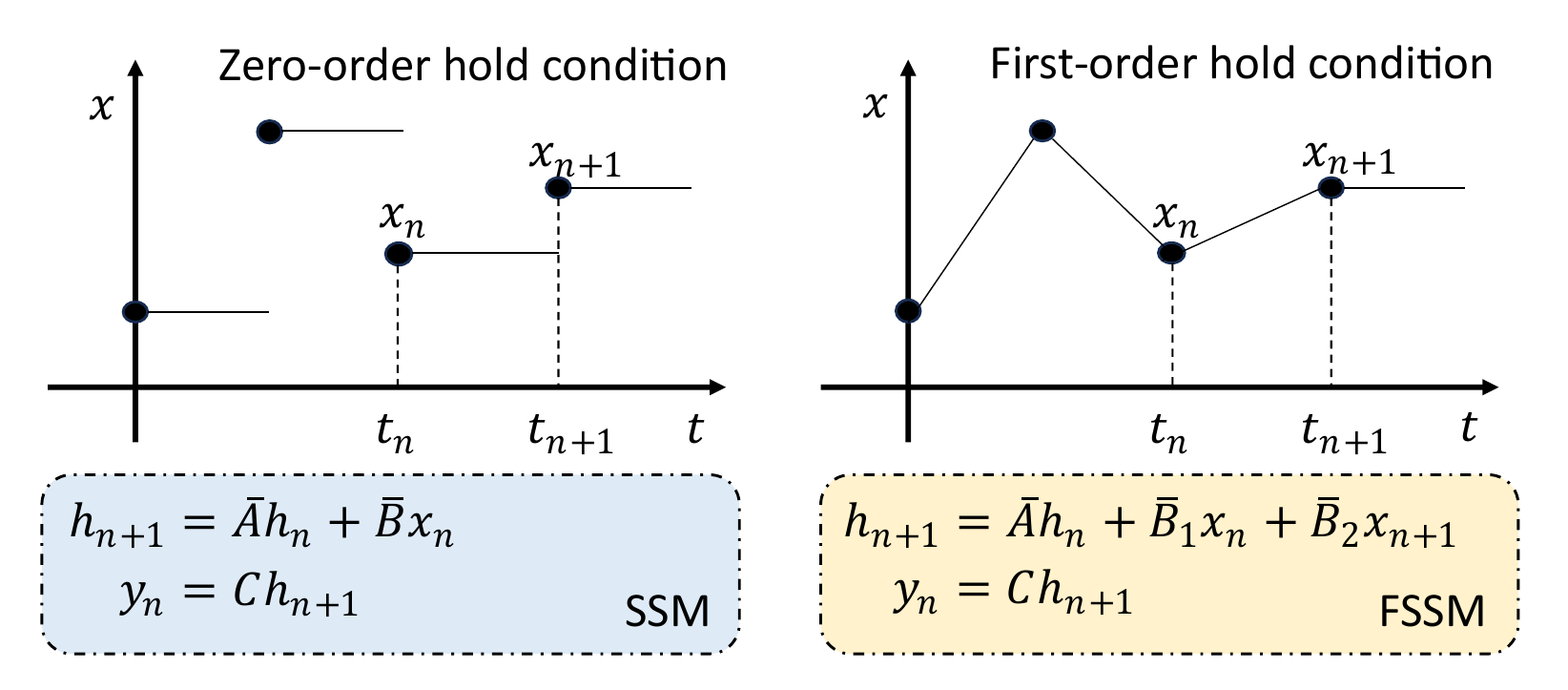}
\end{center}
\vspace{-\baselineskip}
\caption{SSM with different extensions.}
\label{continuous}
\vspace{-\baselineskip}
\end{figure}

For time-invariant SSM, the general solutions of first-order linear ODEs on the interval $[t_n,t_{n+1}]$ can be solved as:
\begin{equation}
\label{fulleq}
    h(t)=e^{(t-t_n)A}h(t_n)+\int_{t_n}^te^{(t-s)A}Bx(s)ds
\end{equation}
where $e^{xA}$ is matrix exponential function, defined as:
\begin{equation}
\label{expeq}
    e^{xA}=I+xA+\frac{(xA)^2}{2!}+\dots+\frac{(xA)^k}{k!}+\dots
\end{equation}
Noticed that traditional SSM usually expand $\{x_n\}$ with zero-order hold condition, which makes that only one-sided information can be used when extracting features by calculating SSM. In order to explore the potential of SSM in vision tasks, and to take full advantage of the relationship between the time steps, we try to improve the model based on the first-order hold condition, Fig.~\ref{continuous} compares the difference between the two extensions.

Based on first-order hold condition, the formula of $x(t)$ over $[t_n,t_{n+1}]$ can be written by:
\begin{equation}
\label{firstlinear}
    x(t)=x_n+\frac{t-t_n}{\Delta_n}(x_{n+1}-x_n)
\end{equation}
where $\Delta_n=t_{n+1}-t_n$. Taking Eq.\eqref{firstlinear} into Eq.\eqref{fulleq}, we can get
\begin{equation*}
\begin{aligned}
    h(t_{n+1})&=e^{\Delta_nA}h(t_n)+\int_{t_n}^{t_{n+1}}e^{(t_{n+1}-s)A}Bx(s)ds \\
    &=e^{\Delta_nA}h_n-A^{-1}\bigg[e^{(t_{n+1}-s)A}Bx(s)\big|_{t_n}^{t_{n+1}} \\
    &\quad-\int_{t_n}^{t_{n+1}}e^{(t_{n+1}-s)A}B\frac{1}{\Delta_n}(x_{n+1}-x_n)ds \bigg] \\
    &=e^{\Delta_nA}h_n-A^{-1}\bigg[Bx_{n+1}-e^{\Delta_nA}Bx_n\\
    &\quad+(\Delta_nA)^{-1}(I-e^{\Delta_nA})B(x_{n+1}-x_n)   \bigg] \\
    &=e^{\Delta_nA}h(t_n) \\
    &\quad+(\Delta_nA)^{-2}(\Delta_nAe^{\Delta_nA}-e^{\Delta_nA}+I)\Delta_nBx_{n} \\
    &\quad+(\Delta_nA)^{-2}(e^{\Delta_nA}-I-\Delta_nA)\Delta_nBx_{n+1}
\end{aligned}
\end{equation*}
Finally, taking $h_n=h(t_n)$ and $h(0)=0$, we get the discrete FSSM form
\begin{equation}
\begin{aligned}
    h_{n+1}&=\overline{A}h_n+\overline{B}_1x_n+\overline{B}_2x_{n+1} \\
    y_n&=Ch_{n+1}  \\
    h_0&=0         \\
    \overline{A}&=e^{\Delta_n A} \\
    \overline{B}_1&=(\Delta_n A)^{-2}(\Delta_n Ae^{\Delta_n A}-e^{\Delta_n A}+I)\Delta_n B \\
    \overline{B}_2&=(\Delta_n A)^{-2}(e^{\Delta_n A}-I-\Delta_n A)\Delta_n B
\end{aligned}
\label{fmamba}
\end{equation}
Similar to Mamba, we incorporate the selection mechanism into models by letting parameters $(\Delta, B, C)$ be time-dependent. We derive $(\Delta_n, B_n, C_n)$ from input $x_n$, and compute the discrete $\overline{A},\overline{B}_1,\overline{B}_2$. Algorithm \ref{alg} provides the pseudo-code of FSSM pipeline.

\begin{algorithm}
\caption{First-order State Space Model (FSSM)}
\label{alg}
\begin{algorithmic}[1]
    \INPUT $A$, $\{x_n\}$ \\
    \OUTPUT $\{y_n\}$ \\
    \STATE $h_0=0$ 
    \WHILE{$n \leq T$}
    \STATE $\Delta_n, B_n, C_n \leftarrow$ linear($x_n$)
    \STATE $\overline{A} \leftarrow$ discretize($\Delta_n, A$)
    \IF{$n \leq T-1$}
    \STATE $\overline{B}_1, \overline{B}_2 \leftarrow$ discretize($\Delta_n,B_n$) \quad \%use Eq.\eqref{fmamba}
    \STATE $h_{n+1}=\overline{A}h_n+\overline{B}_1x_n+\overline{B}_2x_{n+1}$
    \ELSE
    \STATE $\overline{B}\leftarrow$ discretize($\Delta_n,B_n$) \quad \%use Mamba\cite{gu2023mamba} form
    \STATE $h_{n+1} = \overline{A}h_n+\overline{B}x_n$
    \ENDIF
    \STATE $y_n=C_nh_{n+1}$
    \ENDWHILE
    \RETURN $\{y_n\}$
\end{algorithmic}
\end{algorithm}

\subsection{Approximation}

Noticed that the discrete parameter $\overline{B}_1,\overline{B}_2$ has the inverse of $\Delta A$, in order to avoid computing the inverse, we refine the approximation by using the series representation of matrix exponential function Eq.\eqref{expeq}:


\begin{equation}
\label{fssm1}
\begin{aligned}
    \overline{B}_1=\frac12\Delta_n B_n ,\;\overline{B}_2=\frac12\Delta_n B_n
\end{aligned}
\end{equation}
\begin{equation}
\label{fssm2}
\begin{aligned}
    \overline{B}_1=(\frac12+\frac13\Delta_nA)\Delta_n B_n ,\;\overline{B}_2=(\frac12+\frac16\Delta_nA)\Delta_n B_n 
\end{aligned}
\end{equation}

For higher precision, we tried two forms of approximation, and analyzed them with extensive experiments. We called form~\eqref{fssm1} FSSM and form~\eqref{fssm2} FSSM$^+$. We also extend the discrete parameter $\overline{B}$ in Mamba as SSM$^+$
\begin{equation}
\begin{aligned}
\label{mambafull}
    \overline{B}=(1+\frac12\Delta_nA)\Delta_nB_n. 
\end{aligned}
\end{equation}

\subsection{Analysis}
Since we only have discrete inputs $\{x_n\}$ instead of continuous $x(t)$, the missing part contributes to the accumulation of error in the output. Now we analyze the cumulative error of SSM and FSSM, in order to simplify the analysis, we disregard the selective mechanism and assume that ($\Delta,A,B,C$) are time-independent. We consider $\{x_n\}$ are sampled from unknown real $x(t)$, $y(t)$ is solved with $x(t)$ by Eq.\eqref{ssm1}, and $\{y_n\}$ are calculated from SSM(or FSSM) with $\{x_n\}$.

\begin{figure}[hb]
\vspace{-\baselineskip}
\begin{center}
\includegraphics[width=0.98\linewidth]{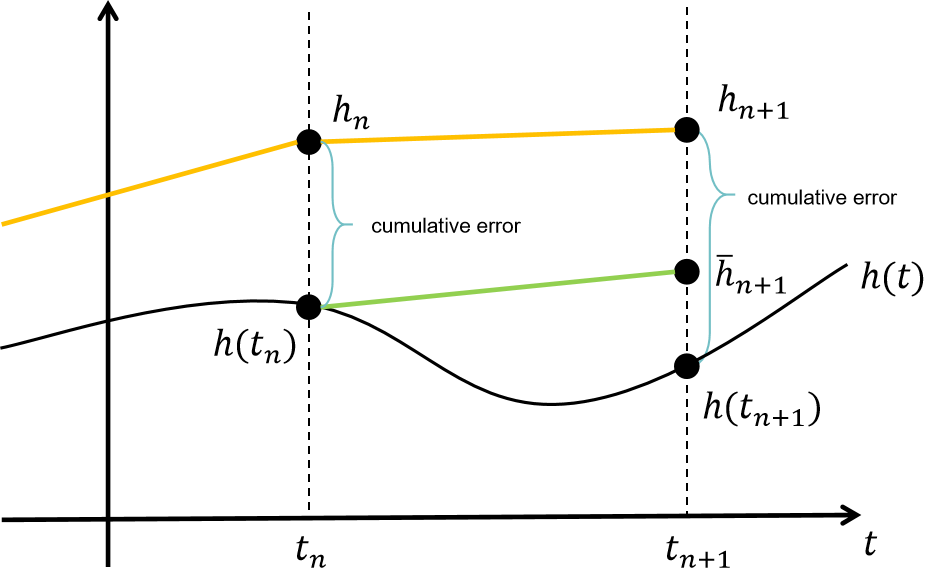}
\end{center}
\vspace{-\baselineskip}
\caption{Components of cumulative error.}
\label{error}
\label{analysis}
\vspace{-\baselineskip}
\end{figure}

\paragraph{Theorem 1}(Cumulative error of SSM). 
Consider discrete SSM form (Mamba\cite{gu2023mamba}) with time-independent $(\Delta, A, B, C)$.
Assume $x(t): \mathbb{R}\rightarrow\mathbb{R}$ is L-Lipschitz and satisfies $x(t_n)=x_n$. There exists $\xi\in(0,\Delta A)$, such that SSM satisfies

\begin{equation}
\begin{aligned}
    |y(t_{n+1})-y_n|\leq C\Delta^2L\frac{e^{(n+1)\Delta A}-I}{e^{\Delta A}-I}e^{\xi}|B|.
\end{aligned}
\end{equation}

\paragraph{Theorem 2}(Cumulative error of FSSM). 
Consider discrete FSSM form~\eqref{fmamba} with time-independent $(\Delta, A, B, C)$.
Assume $x(t): \mathbb{R}\rightarrow\mathbb{R}$ is L-Lipschitz and satisfies $x(t_n)=x_n$. There exists $\xi\in(0,\Delta A)$, such that FSSM satisfies

\begin{equation}
\begin{aligned}
    |y(t_{n+1})-y_n|\leq \frac{C\Delta^2L}{2}\frac{e^{(n+1)\Delta A}-I}{e^{\Delta A}-I}e^{\xi}|B|.
\end{aligned}
\end{equation}


See Appendix \ref{proofs} for complete proof details. Noticed that FSSM has the same order of error as SSM, but has a lower upper bound of cumulative error.

\section{Experiments}
\label{experiments}

\subsection{Datasets and Details}
\label{details}
\paragraph{Datasets and Evaluation} We train our SR models on DIV2K~\cite{agustsson2017ntire} and DF2K(DIV2K + Flickr2K~\cite{lim2017enhanced}) datasets. Following other SR works, we use Set5~\cite{bevilacqua2012low}, Set14~\cite{zeyde2012single}, B100~\cite{martin2001database}, Urban100~\cite{huang2015single}, and Manga109~\cite{matsui2017sketch} to evaluate the generalizability and effectiveness of different SR models. For quantitative metrics, we use PSNR and SSIM\cite{wang2004image} scores calculated on the Y channel of the YCbCr space. 

\paragraph{Training Details} According to previous works of SR~\cite{chen2023activating,liang2021swinir,guo2024mambair}, we perform data augmentation by applying horizontal flips and random rotations of 90$^\circ$, 180$^\circ$, and 270$^\circ$, as well as cropping the original images into 64×64 patches. For network and training process, we choose identical network architecture and training settings with MambaIR. Additionally, we use pretrained weights from MambaIR as an initialization of corresponding upscaling ratio. Our model is trained with 4 NVIDIA 3090 GPUs.


\begin{table*}[]
\caption{Quantitative comparison with state-of-the-art methods on benchmark datasets.The best results are highlighted in black bold and the second best is in underline.}
\centering
\scalebox{0.85}{
\begin{tabular}{c|c|c|c|c|c|c|c|c}
\hline
\multirow{2}{*}{Method} & \multirow{2}{*}{Scale} & \multirow{2}{*}{Params} & Training & \multicolumn{1}{c|}{Set5} & \multicolumn{1}{c|}{Set14} & \multicolumn{1}{c|}{B100} & \multicolumn{1}{c|}{Urban100} & \multicolumn{1}{c}{Manga109} \\ 
\cline{5-9} 
                                        &       &           & Dataset   & PSNR / SSIM      & PSNR / SSIM      & PSNR / SSIM        & PSNR / SSIM      & PSNR / SSIM      \\ 
\hline
CARN~\cite{ahn2018fast}                 & \multirow{9}{*}{x2}    & 1,592K    & DIV2K     & 37.76 / 0.9590    & 33.52 / 0.9166    & 32.09 / 0.8978      & 31.92 / 0.9256    & 38.36 / 0.9765    \\
IMDN~\cite{hui2019lightweight}          &     & 694K      & DIV2K     & 38.00 / 0.9605    & 33.63 / 0.9177    & 32.19 / 0.8996      & 32.17 / 0.9283    & 38.88 / 0.9774    \\
LAPAR-A~\cite{li2020lapar}              &     & 548K      & DF2K      & 38.01 / 0.9605    & 33.62 / 0.9183    & 32.19 / 0.8999      & 32.10 / 0.9283    & 38.67 / 0.9772    \\
LatticeNet~\cite{luo2020latticenet}     &     & 756K      & DIV2K     & 38.15 / 0.9610    & 33.78 / 0.9193    & 32.25 / 0.9005      & 32.43 / 0.9302    & - / -         \\
SwinIR-light~\cite{liang2021swinir}     &     & 878K      & DIV2K     & 38.14 / 0.9611    & 33.86 / 0.9206    & 32.31 / 0.9012      & 32.76 / 0.9340    & 39.12 / 0.9783    \\
SRFormer-light~\cite{zhou2023srformer}  &     & 853K      & DIV2K     &\underline{38.23} / 0.9613    & 33.94 / 0.9209    & 32.36 / 0.9019      & 32.91 / 0.9353    & 39.28 / 0.9785    \\
MambaIR-light~\cite{guo2024mambair}     &     & 859K      & DIV2K     & 38.16 / 0.9610    & 34.00 / 0.9212    & 32.34 /0.9017      & 32.92 /0.9356    & 39.31 / 0.9779    \\
\textbf{FMambaIR}(ours)                 &     & 859K      & DIV2K     & 38.20 / \underline{0.9616}    &\underline{34.12} / \underline{0.9225} &\underline{32.37} / \underline{0.9029} &\underline{33.16} / \underline{0.9372} &\underline{39.43} / \underline{0.9788} \\
\textbf{FMambaIR}(ours)                 &     & 859K      & DF2K      &\textbf{38.25} / \textbf{0.9618}  &\textbf{34.21} / \textbf{0.9230} &\textbf{32.41} / \textbf{0.9033} &\textbf{33.29} / \textbf{0.9381} &\textbf{39.59} / \textbf{0.9792} \\ 
\hline
CARN~\cite{ahn2018fast}                 & \multirow{9}{*}{x3}    & 1,592K    & DIV2K     & 34.29 / 0.9255    & 30.29 / 0.8407    & 29.06 / 0.8034      & 28.06 / 0.8493    & 33.50 / 0.9440    \\
IMDN~\cite{hui2019lightweight}          &     & 703K      & DIV2K     & 34.36 / 0.9270    & 30.32 / 0.8417    & 29.09 / 0.8046      & 28.17 / 0.8519    & 33.61 / 0.9445    \\
LAPAR-A~\cite{li2020lapar}              &     & 544K      & DF2K      & 34.36 / 0.9267    & 30.34 / 0.8421    & 29.11 / 0.8054      & 28.15 / 0.8523    & 33.51 / 0.9441    \\
LatticeNet~\cite{luo2020latticenet}     &     & 765K      & DIV2K     & 34.53 / 0.9281    & 30.39 / 0.8424    & 29.15 / 0.8059      & 28.33 / 0.8538    & - / -         \\
SwinIR-light~\cite{liang2021swinir}     &     & 886K      & DIV2K     & 34.62 / 0.9289    & 30.54 / 0.8463    & 29.20 / 0.8082      & 28.66 / 0.8624    & 33.98 / 0.9478    \\
SRFormer-light~\cite{zhou2023srformer}  &     & 861K      & DIV2K     & 34.67 / 0.9296    & 30.57 / 0.8469    & 29.26 / 0.8099      & 28.81 / 0.8655    & 34.19 / 0.9489    \\
MambaIR-light~\cite{guo2024mambair}     &     & 867K      & DIV2K     & 34.72 / 0.9296    & 30.63 / 0.8475    & 29.29 / 0.8099      & 29.00 / 0.8689    & 34.39 / 0.9495    \\
\textbf{FMambaIR}(ours)                 &     & 867K      & DIV2K     &\underline{34.73} / \underline{0.9301} &\underline{30.63} / \underline{0.8482} &\underline{29.31} / \underline{0.8119} &\underline{29.02} / \underline{0.8690} &\underline{34.39} / \underline{0.9497}    \\
\textbf{FMambaIR}(ours)                 &     & 867K      & DF2K      &\textbf{34.78} / \textbf{0.9304} &\textbf{30.65} / \textbf{0.8487} &\textbf{29.33} / \textbf{0.8124} &\textbf{29.13} / \textbf{0.8712} &\textbf{34.65} / \textbf{0.9505} \\ 
\hline
CARN~\cite{ahn2018fast}                 & \multirow{9}{*}{x4}    & 1,592K    & DIV2K     & 32.13 / 0.8937    & 28.60 / 0.7806    & 27.58 / 0.7349      & 26.07 / 0.7837    & 30.47 / 0.9084    \\
IMDN~\cite{hui2019lightweight}          &     & 715K      & DIV2K     & 32.21 / 0.8948    & 28.58 / 0.7811    & 27.56 / 0.7353      & 26.04 / 0.7838    & 30.45 / 0.9075    \\
LAPAR-A~\cite{li2020lapar}              &     & 659K      & DF2K      & 32.15 / 0.8944    & 28.61 / 0.7818    & 27.61 / 0.7366      & 26.14 / 0.7871    & 30.42 / 0.9074    \\
LatticeNet~\cite{luo2020latticenet}     &     & 777K      & DIV2K     & 32.30 / 0.8962    & 28.68 / 0.7830    & 27.62 / 0.7367      & 26.25 / 0.7873    & - / -         \\
SwinIR-light~\cite{liang2021swinir}     &     & 897K      & DIV2K     & 32.44 / 0.8976    & 28.77 / 0.7858    & 27.69 / 0.7406      & 26.47 / 0.7980    & 30.92 / 0.9151    \\
SRFormer-light~\cite{zhou2023srformer}  &     & 873K      & DIV2K     & 32.51 / 0.8988    & 28.82 / 0.7872    & 27.73 / 0.7422      & 26.67 / 0.8032    & 31.17 /0.9165    \\
MambaIR-light~\cite{guo2024mambair}     &     & 879K      & DIV2K     & 32.51 / 0.8993    &\underline{28.85} / 0.7876    & 27.75 / 0.7423      & 26.75 / 0.8051    &\underline{31.26} / \underline{0.9175}    \\
\textbf{FMambaIR}(ours)                 &     & 879K      & DIV2K     &\underline{32.51} / \underline{0.9000}    & 28.83 / \underline{0.7878}    &\underline{27.77} / \underline{0.7448} & \underline{26.78} / \underline{0.8053}  & 31.16 / 0.9169    \\
\textbf{FMambaIR}(ours)                 &     & 879K      & DF2K      &\textbf{32.56} / \textbf{0.9002} &\textbf{28.94} / \textbf{0.7896} &\textbf{27.80} / \textbf{0.7457} &\textbf{26.83} / \textbf{0.8076} &\textbf{31.42} / \textbf{0.9187} \\ 
\hline
\end{tabular}}
\vspace{-\baselineskip}
\label{compare_table}
\end{table*}

\subsection{Experiment results}
\begin{figure}[]
	\centering
	\begin{minipage}{0.45\linewidth}
		\centering
		\includegraphics[width=\linewidth]{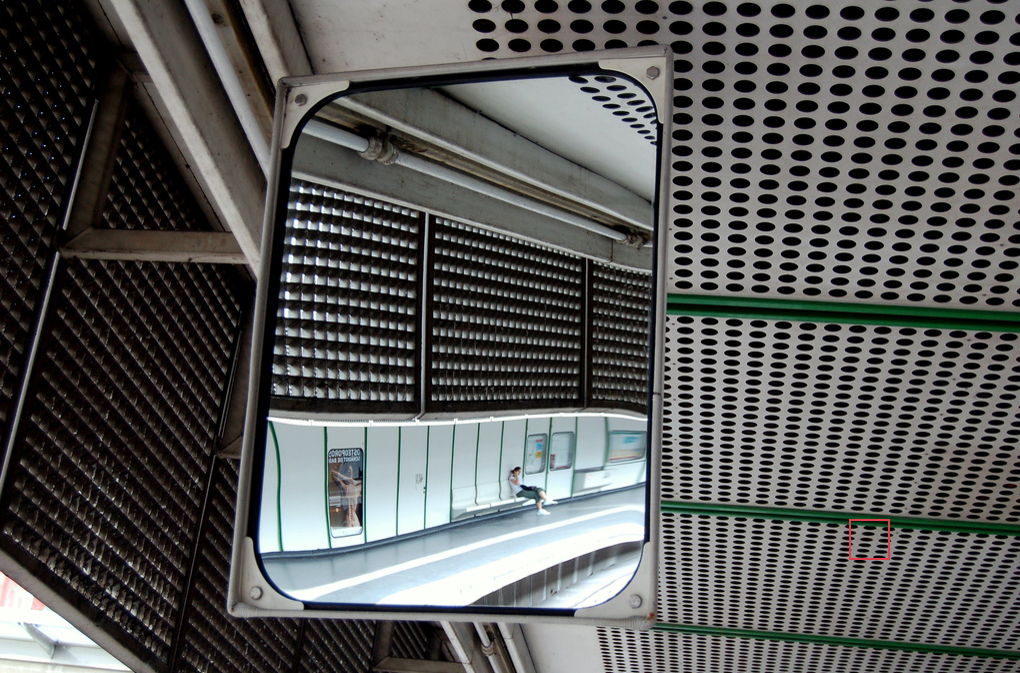}
	\end{minipage}
	\begin{minipage}{0.53\linewidth}
		\centering
		\includegraphics[width=\linewidth]{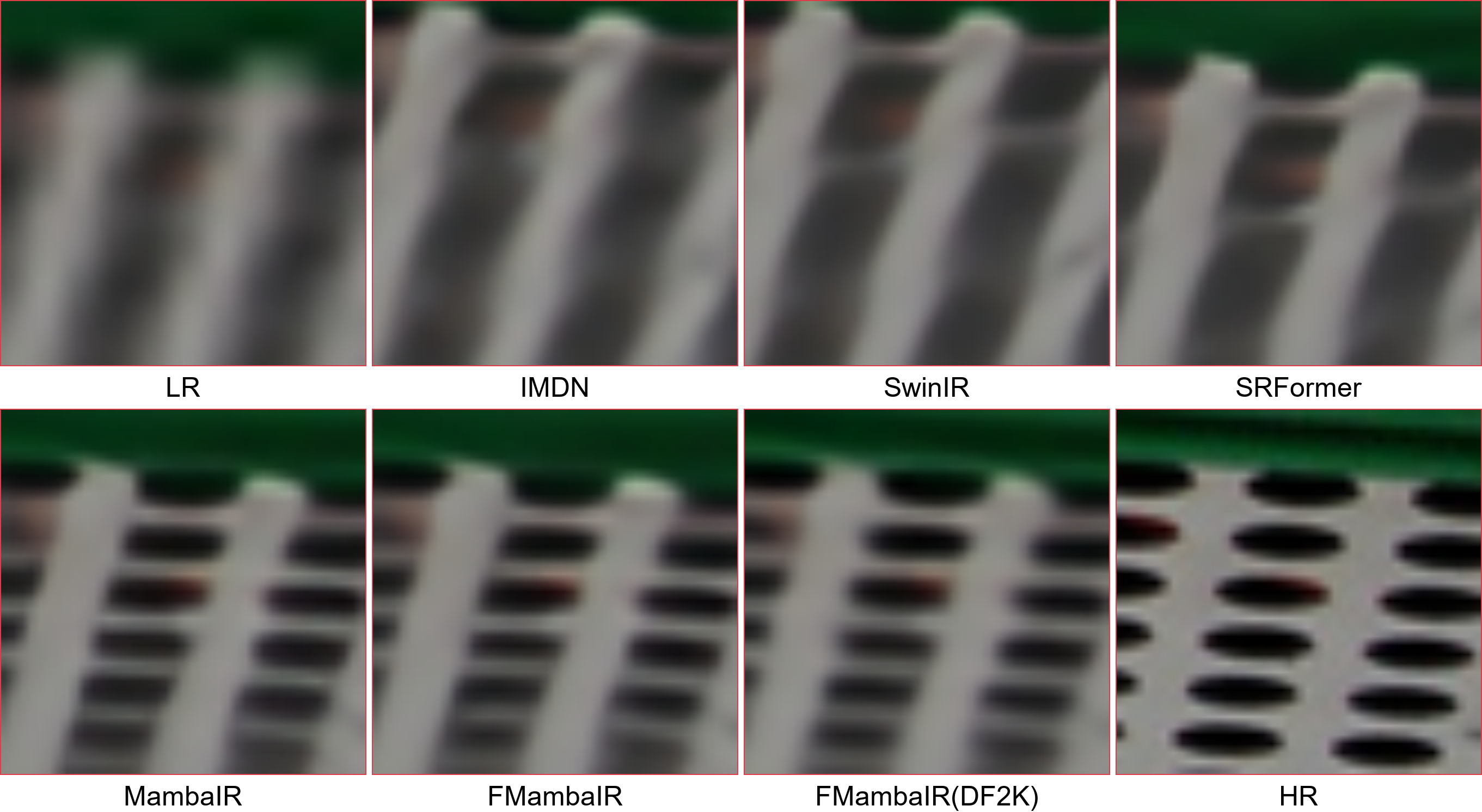}
	\end{minipage}
	
	\begin{minipage}{0.45\linewidth}
		\centering
		\includegraphics[width=\linewidth]{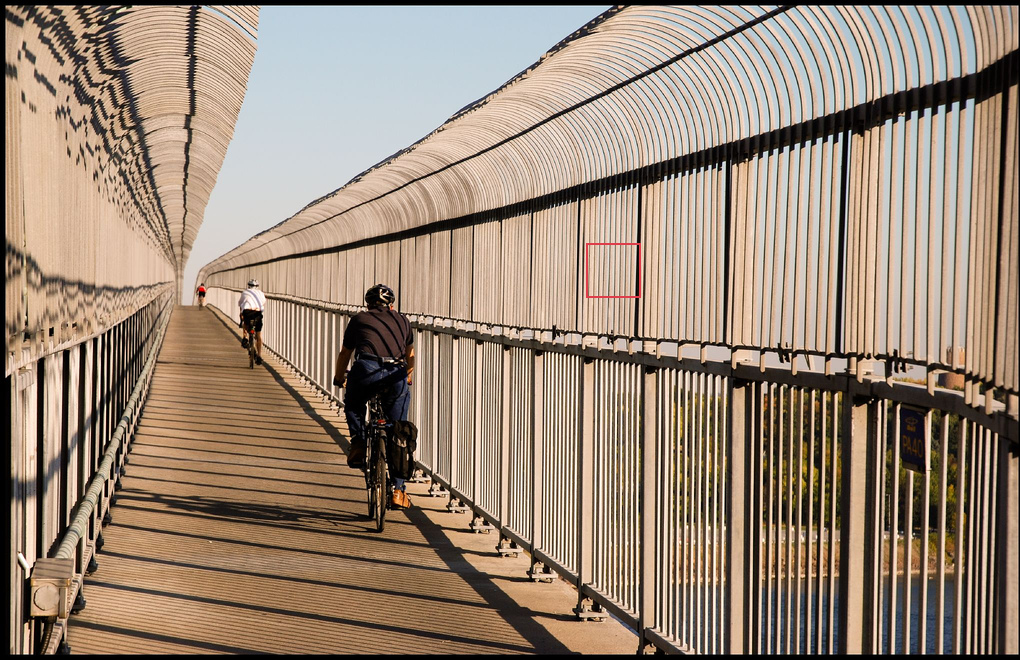}
	\end{minipage}
	\begin{minipage}{0.53\linewidth}
		\centering
		\includegraphics[width=\linewidth]{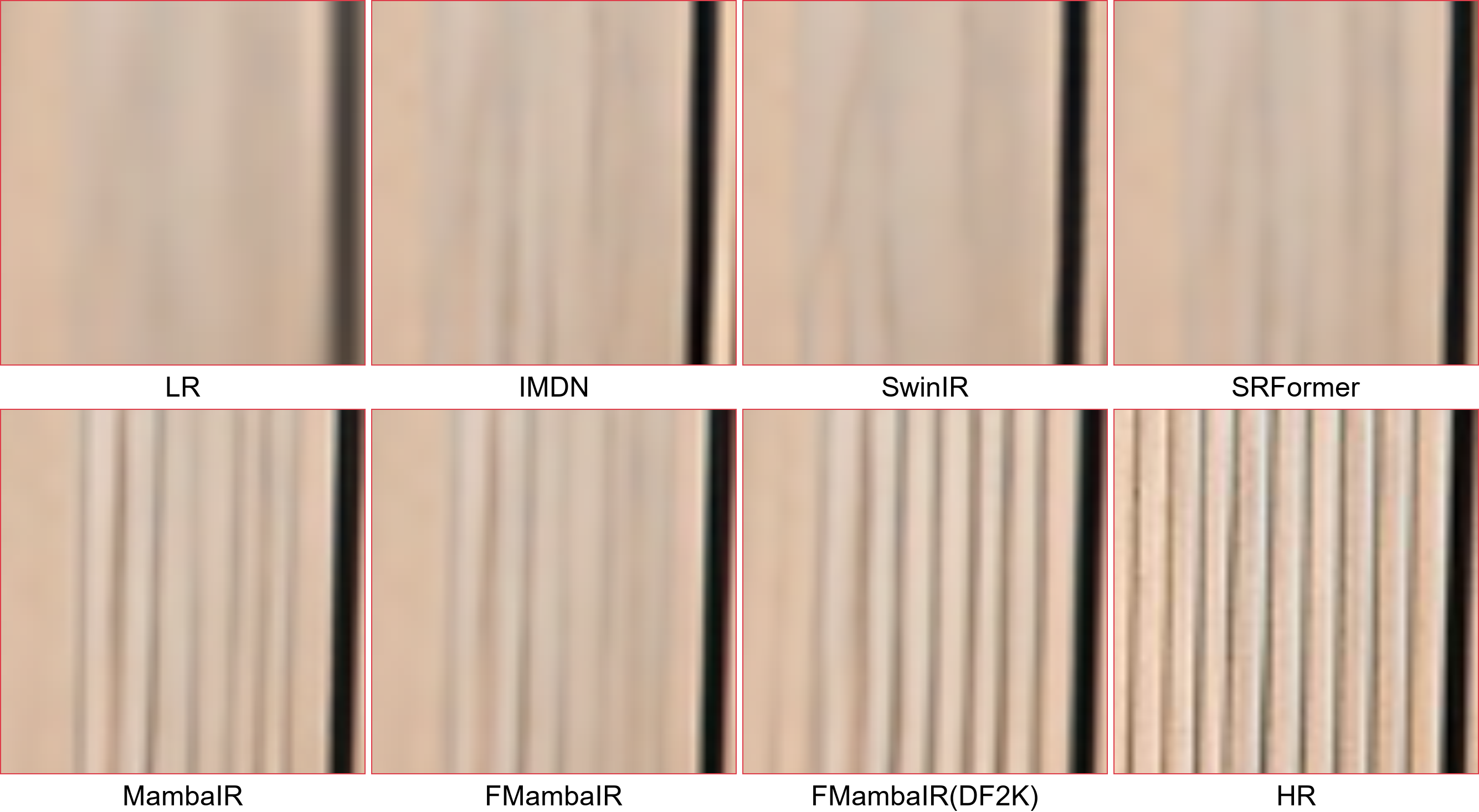}
	\end{minipage}
     \caption{Qualitative Comparison of our FMambaIR model with SOTA lightweight SR models on x4 upscaling task.}
     \label{images}
\vspace{-\baselineskip}
\end{figure}

We compare our MambaIR + FSSM$^+$ (FMambaIR) model with several other state-of-the-art light-weight SR methods, including CARN\cite{ahn2018fast}, IMDN\cite{hui2019lightweight}, LAPAR-A\cite{li2020lapar}, LatticeNet\cite{luo2020latticenet}, SwinIR-light\cite{liang2021swinir}, SRFormer-light\cite{zhou2023srformer} and MambaIR-light\cite{guo2024mambair}. Table~\ref{compare_table} demonstrates that our proposed FMambaIR outperforms other methods almost on all five benchmark datasets and all scales. Specifically, our FMambaIR model surpasses original MambaIR by up to 0.24dB PSNR on the x2 scale Urban100 dataset with the same network architecture and amount of parameters. Meanwhile, visual results comparison is provided in Fig~\ref{images}. We observe that our model recover images more accurately while other approaches introduce some obvious structural distortions into recovered results.

\subsection{Ablation Study}
\paragraph{Effects of different extensions}
Expand condition is the core component of SSM. Using higher-order expand condition can improve the correlation between the tokens, and reduce the cumulative error. In this section, we ablate different extensions of SSM. In Table~\ref{compare_table} we compared SSM and FSSM$^+$ module trained under dataset DIV2K, for more comparative experiments, we also trained two methods under DF2K dataset. The PSNR and SSIM results of x2, x3, x4 light-weight SR presented in Table~\ref{compare_rule} show that FSSM$^+$ module can enhance the performance since we take the correlation between tokens into account.

\begin{table}[]
\vspace{-\baselineskip}
\caption{Effects of different extensions of SSM on benchmark datasets.}
\label{compare_rule}
\centering
\scalebox{0.67}{
\begin{tabular}{c|c|c|c|c|c|c}
\hline
\multirow{2}{*}{Module} & \multirow{2}{*}{Scale} & \multicolumn{1}{c|}{Set5} & \multicolumn{1}{c|}{Set14} & \multicolumn{1}{c|}{B100} & \multicolumn{1}{c|}{Urban100} & \multicolumn{1}{c}{Manga109} \\ \cline{3-7} 
        &       & PSNR / SSIM      & PSNR / SSIM      & PSNR / SSIM        & PSNR / SSIM      & PSNR / SSIM \\ 
        \hline
SSM   & \multirow{2}{*}{x2}   & 38.23 / 0.9618  & 34.17 / 0.9226  & 32.38 / 0.9030  & 33.19 / 0.9377 & 39.49 / 0.9789 \\ 
FSSM$^+$   &                  & 38.25 / 0.9618  & 34.21 / 0.9230  & 32.41 / 0.9033  & 33.29 / 0.9381 & 39.59 / 0.9792 \\
\hline
SSM   & \multirow{2}{*}{x3}   & 34.76 / 0.9302  & 30.65 / 0.8484  & 29.33 / 0.8123  & 29.09 / 0.8701 & 34.51 / 0.9500 \\
FSSM$^+$   &                  & 34.78 / 0.9304  & 30.65 / 0.8487  & 29.33 / 0.8124  & 29.13 / 0.8712 & 34.65 / 0.9505 \\
\hline
SSM   & \multirow{2}{*}{x4}   & 32.54 / 0.9000  & 28.92 / 0.7893  & 27.78 / 0.7450  & 26.82 / 0.8074 & 31.38 / 0.9184 \\
FSSM$^+$   &                  & 32.56 / 0.9002  & 28.94 / 0.7896  & 27.80 / 0.7457  & 26.83 / 0.8076 & 31.42 / 0.9187 \\
\hline
\end{tabular}}
\vspace{-\baselineskip}
\end{table}

\paragraph{Effects of different discretization approximation}
Increasing the expansion order can reduce the truncation error, and thus reduce the cumulative error. In this section, we compare different expansion order of approximation,  FSSM (Eq.\eqref{fssm1}) and FSSM$^+$ (Eq.\eqref{fssm2}) on benchmark datasets. The PSNR results of x2, x3, x4 light-weight SR presented in Table~\ref{compare_expand} show that more expansion terms can improve the performance. Note that with higher expansion, the remainder contains less valid information, so the improvement becomes smaller.

\begin{table}[]
\caption{Effects of different discretization approximation of SSM on benchmark datasets.}
\centering
\scalebox{0.67}{
\begin{tabular}{c|c|c|c|c|c|c}
\hline
\multirow{2}{*}{Settings} & \multirow{2}{*}{Scale} & \multicolumn{1}{c|}{Set5} & \multicolumn{1}{c|}{Set14} & \multicolumn{1}{c|}{B100} & \multicolumn{1}{c|}{Urban100} & \multicolumn{1}{c}{Manga109} \\ \cline{3-7} 
        &       & PSNR / SSIM      & PSNR / SSIM      & PSNR / SSIM        & PSNR / SSIM      & PSNR / SSIM \\ 
\hline
FSSM    & \multirow{2}{*}{x2}  & 38.24 / 0.9618  & 34.21 / 0.9230  & 32.40 / 0.9033  & 33.29 / 0.9381 & 39.59 / 0.9792 \\
FSSM$^+$      &                & 38.25 / 0.9618  & 34.21 / 0.9230  & 32.41 / 0.9033  & 33.29 / 0.9381 & 39.59 / 0.9792 \\ 
\hline
FSSM    & \multirow{2}{*}{x3}  & 34.75 / 0.9302  & 30.65 / 0.8485  & 29.32 / 0.8122  & 29.12 / 0.8710 & 34.63 / 0.9505 \\
FSSM$^+$      &                & 34.78 / 0.9304  & 30.65 / 0.8487  & 29.33 / 0.8124  & 29.13 / 0.8712 & 34.65 / 0.9505 \\ 
\hline
FSSM    & \multirow{2}{*}{x4}  & 32.55 / 0.9002  & 28.89 / 0.7885  & 27.78 / 0.7447  & 26.81 / 0.8067 & 31.35 / 0.9180 \\
FSSM$^+$      &                & 32.56 / 0.9002  & 28.94 / 0.7896  & 27.80 / 0.7457  & 26.83 / 0.8076 & 31.42 / 0.9187 \\
\hline
\end{tabular}}
\label{compare_expand}
\vspace{-\baselineskip}
\end{table}


\section{Conclusion}

In this paper, we propose the First-order State Space Model (FSSM) to enhance the fundamental SSM module in Mamba architecture. We derive a new discretization form based on the first-order hold condition, use higher approximation to enhance the computational precision and analyze the cumulative error. Experiments show that FSSM improves the basic Mamba module and outperforms state-of-the-art methods in lightweight super-resolution. Given its promising results compared to vanilla Mamba, we are excited about FSSM's potential across various tasks and domains.



\bibliographystyle{IEEEtran}
\bibliography{IEEEabrv, citation}

\appendix


\section{Analysis of cumulative error}
\label{proofs}
\subsection{Mamba}
\textbf{Theorem 1}(Cumulative error of Mamba). 
Consider discrete SSM form (Mamba\cite{gu2023mamba}) with time-independent $(\Delta, A, B, C)$.
\begin{equation}
\begin{aligned}
\label{mambasimple}
    h_{n+1}&=e^{\Delta A}h_n+\overline Bx_n \\
    y_n&=Ch_{n+1}  \\
    h_0&=0 \\
    \overline B &=(\Delta A)^{-1}(e^{\Delta A}-I)\Delta B
\end{aligned}
\end{equation}
Assume $x(t): \mathbb{R}\rightarrow\mathbb{R}$ is L-Lipschitz and satisfies $x(t_n)=x_n$. There exists $\xi\in(0,\Delta A)$, such that SSM satisfies

\begin{equation}
\begin{aligned}
\label{theoroy1}
    |y(t_{n+1})-y_n|\leq C\Delta^2L\frac{e^{(n+1)\Delta A}-I}{e^{\Delta A}-I}e^{\xi}|B|.
\end{aligned}
\end{equation}

\emph{Proof.}
Let $h(t)$ be the global solution of SSM with boundary condition $h(0)=0$, define $\overline{h}_{n+1}$ as the solution of SSM at $t_{n+1}$ with zero-order hold condition 
\begin{equation*}
\begin{aligned}
    h(t_{n+1})&=e^{\Delta A}h(t_n)+\int_{t_n}^{t_{n+1}}e^{(t_{n+1}-s)A}Bx(s)ds \\
    \overline{h}_{n+1}
    &=e^{\Delta A}h(t_n)+\int_{t_n}^{t_{n+1}}e^{(t_{n+1}-s)A}B\overline{x}(s)ds
\end{aligned}
\end{equation*}

Noticed that $\overline{x}(t)$ holds $x_n$ on $[t_n,t_{n+1})$, and $x(t)$ is L-Lipschitz, then 
\begin{equation*}
    |x(t)-\overline{x}(t)|\leq \Delta L 
\end{equation*}
holds on $[t_n,t_{n+1})$. Then
\begin{equation}
\begin{aligned}
    |h(t_{n+1})-\overline{h}_{n+1}|&\leq\int_{t_n}^{t_{n+1}}e^{(t_{n+1}-s)A}|B||x(s)-\overline{x}(s)|ds \\
    &\leq \Delta L\int_{t_n}^{t_{n+1}}e^{(t_{n+1}-s)A}|B|ds \\
    &=\Delta LA^{-1}(e^{\Delta A}-I)|B| \\
    &=\Delta^2Le^{\xi}|B|\quad \xi\in(0,\Delta A)
\end{aligned}
\end{equation}

Let $C_1=Le^{\xi}|B|$. Noticed that $\overline{h}_{n+1}$ have explicit form
\begin{equation*}
    \overline{h}_{n+1}=e^{\Delta A}h(t_n)+\overline{B}x_n.
\end{equation*}

And $h_{n+1}$ is the discrete approximation of $h(t_{n+1})$, with boundary condition $h(t_n)=h_n$ as Eq.\eqref{mambasimple}, then
\begin{equation}
    |\overline{h}_{n+1}-h_{n+1}|= e^{\Delta A}|h(t_n)-h_n|
\end{equation}

Then cumulative error at $t_{n+1}$ is
\begin{equation}
\begin{aligned}
\label{theory1end}
    |h(t_{n+1})-h_{n+1}|&\leq |h(t_{n+1})-\overline{h}_{n+1}|+|\overline{h}_{n+1}-h_{n+1}| \\
    &\leq e^{\Delta A}|h(t_n)-h_n|+\Delta^2C_1 \\
    |h(t_{n+1})-h_{n+1}|+\frac{\Delta^2C_1}{e^{\Delta A}-I}&\leq e^{\Delta A}\left(|h(t_n)-h_n|+\frac{\Delta^2C_1}{e^{\Delta A}-I}\right) \\
    |h(t_{n+1})-h_{n+1}|&\leq (e^{(n+1)\Delta A}-I)\frac{\Delta^2C_1}{e^{\Delta A}-I}.
\end{aligned}
\end{equation}
Substituting Eq.\eqref{theory1end} into Eq.\eqref{mambasimple} yields Eq.\eqref{theoroy1}. \quad$\Box$

\subsection{FMamba}
\textbf{Theorem 2}(Cumulative error of FMamba). 
Consider discrete FSSM form~\eqref{fmamba} with time-independent $(\Delta, A, B, C)$.
\begin{equation}
\begin{aligned}
\label{fmambasimple}
    h_{n+1}&=e^{\Delta A}h_n+\overline B_1x_n+\overline B_2x_{n+1} \\
    y_n&=Ch_{n+1}  \\
    h_0&=0 \\
    \overline{B}_1&=(\Delta A)^{-2}(\Delta Ae^{\Delta A}-e^{\Delta A}+I)\Delta B \\
    \overline{B}_2&=(\Delta A)^{-2}(e^{\Delta A}-I-\Delta A)\Delta B
\end{aligned}
\end{equation}
Assume $x(t): \mathbb{R}\rightarrow\mathbb{R}$ is L-Lipschitz and satisfies $x(t_n)=x_n$. There exists $\xi\in(0,\Delta A)$, such that FSSM satisfies
\begin{equation}
\begin{aligned}
\label{theoroy2}
    |y(t_{n+1})-y_n|\leq \frac{C\Delta^2L}{2}\frac{e^{(n+1)\Delta A}-I}{e^{\Delta A}-I}e^{\xi}|B|.   
\end{aligned}
\end{equation}

\emph{Proof.}
Let $h(t)$ be the global solution of SSM with boundary condition $h(0)=0$, define $\overline{h}_{n+1}$ as the solution of SSM at $t_{n+1}$ with first-order hold condition 

\begin{equation*}
\begin{aligned}
    h(t_{n+1})&=e^{\Delta A}h(t_n)+\int_{t_n}^{t_{n+1}}e^{(t_{n+1}-s)A}Bx(s)ds \\
    \overline{h}_{n+1}&=e^{\Delta A}h(t_n)+\int_{t_n}^{t_{n+1}}e^{(t_{n+1}-s)A}B\overline{x}(s)ds
\end{aligned}
\end{equation*}

The formula of $\overline{x}(t)$ over $[t_n,t_{n+1}]$ is
\begin{equation*}
    \overline x(t)=x_n+\frac{t-t_n}{\Delta_n}(x_{n+1}-x_n)
\end{equation*}

Noticed that $x(t)$ is L-Lipschitz, then 
\begin{equation*}
    |x(t)-\overline{x}(t)|\leq \Delta L/2 
\end{equation*}
holds on $[t_n,t_{n+1}]$. Then
\begin{equation}
\begin{aligned}
    |h(t_{n+1})-\overline{h}_{n+1}|&\leq\int_{t_n}^{t_{n+1}}e^{(t_{n+1}-s)A}|B||x(s)-\overline{x}(s)|ds \\
    &\leq \frac12\Delta L\int_{t_n}^{t_{n+1}}e^{(t_{n+1}-s)A}|B|ds \\
    &=\frac12\Delta LA^{-1}(e^{\Delta A}-I)|B| \\
    &=\Delta^2Le^{\xi}|B|/2\quad \xi\in(0,\Delta A)
\end{aligned}
\end{equation}

Let $C_1=Le^{\xi}|B|/2$. Noticed that $\overline{h}_{n+1}$ have explicit form
\begin{equation*}
    \overline{h}_{n+1}=e^{\Delta A}h(t_n)+\overline{B}_1x_n + \overline{B}_2 x_{n+1}.
\end{equation*}
And $h_{n+1}$ is the discrete approximation of $h(t_{n+1})$, with boundary condition $h(t_n)=h_n$, as Eq.\eqref{fmambasimple}, then
\begin{equation}
\begin{aligned}
    |\overline{h}_{n+1}-h_{n+1}|= e^{\Delta A}|h(t_n)-h_n|
\end{aligned}
\end{equation}

Then cumulative error at $t_{n+1}$ is
\begin{equation}
\begin{aligned}
\label{theory2end}
    |h(t_{n+1})-h_{n+1}|&\leq |h(t_{n+1})-\overline{h}_{n+1}|+|\overline{h}_{n+1}-h_{n+1}| \\
    &\leq e^{\Delta A}|h(t_n)-h_n|+\Delta^2C_1 \\
    |h(t_{n+1})-h_{n+1}|+\frac{\Delta^2C_1}{e^{\Delta A}-I}&\leq e^{\Delta A}\left(|h(t_n)-h_n|+\frac{\Delta^2C_1}{e^{\Delta A}-I}\right) \\
    |h(t_{n+1})-h_{n+1}|&\leq (e^{(n+1)\Delta A}-I)\frac{\Delta^2C_1}{e^{\Delta A}-I}.
\end{aligned}
\end{equation}
Substituting Eq.\eqref{theory2end} into Eq.\eqref{fmambasimple} yields Eq.\eqref{theoroy2}. \quad$\Box$

\end{document}